\documentclass{article}

\usepackage{times}
\usepackage{graphicx} 
\usepackage{subfigure,overpic} 

\usepackage{natbib}

\usepackage{algorithm}
\usepackage{algorithmic}

\usepackage{amsfonts}
\usepackage{amsmath}
\usepackage{amssymb}
\usepackage{amsthm}

\usepackage{tikz}
\usepackage[utf8]{inputenc}
\usepackage{pgfplots}
\pgfplotsset{compat=newest}
\usepackage{hyperref}

\usepackage[accepted]{icml2017}

\icmltitlerunning{Geometric Matrix Completion with Recurrent Multi-Graph Neural Networks}

\pgfplotsset{every axis/.append style={
                    label style={font=\tiny},
                    tick label style={font=\tiny}  
                    }}
                    
\usepgfplotslibrary{groupplots}                 

\newlength\figureheight
\newlength\figurewidth

\newcommand{\etal}{\textit{et al}. }

\begin{document} 

\twocolumn[
\icmltitle{Geometric Matrix Completion with Recurrent Multi-Graph Neural Networks}



\icmlsetsymbol{equal}{*}

\begin{icmlauthorlist}
\icmlauthor{Federico Monti}{usi}
\icmlauthor{Michael M. Bronstein}{usi,tau,intel,tum}
\icmlauthor{Xavier Bresson}{ntu}
\end{icmlauthorlist}

\icmlaffiliation{usi}{ICS USI Lugano, Switzerland}
\icmlaffiliation{tau}{Tel Aviv University, Israel}
\icmlaffiliation{intel}{Intel Perceptual Computing, Israel}
\icmlaffiliation{ntu}{NTU, Singapore}
\icmlaffiliation{tum}{TUM IAS, Germany}

\icmlcorrespondingauthor{Federico Monti}{federico.monti@usi.ch}

\icmlkeywords{geometric deep learning, matrix completion, recommender systems}

\vskip 0.3in
]



\printAffiliationsAndNotice{}  

\begin{abstract}

Matrix completion models are among the most common formulations of recommender
systems. Recent works have showed a boost of performance of these techniques when introducing the pairwise relationships between users/items in the form of graphs, and imposing smoothness priors on these graphs. However, such techniques do not fully exploit the local stationarity structures of user/item graphs, and the number of parameters to learn is linear w.r.t. the number of users and items. We propose a novel approach to overcome these limitations by using geometric deep learning on graphs. Our matrix completion architecture combines graph convolutional neural networks and recurrent neural networks to learn meaningful statistical graph-structured patterns and the non-linear diffusion process that generates the known ratings. This neural network system requires a constant number of parameters independent of the matrix size. We apply our method on both synthetic and real datasets, showing that it outperforms state-of-the-art techniques.

\end{abstract} 

\section{Introduction}

Recommender systems have become a central part of modern intelligent systems. Recommending movies on Netflix, friends on Facebook, furniture on Amazon, jobs on LinkedIn are a few examples of the main purpose of these systems. 
Two major approach to recommender systems are collaborative \cite{pro:BreeseHeckermanKadie98CollFilt} and content \cite{art:PazzaniBillsus07ContFilt} filtering techniques. Systems based on collaborative filtering use collected ratings of products by customers and offer new recommendations by finding similar rating patterns. Systems based on content filtering make use of similarities between products and customers to recommend new products. Hybrid systems combine collaborative and content techniques.

\paragraph*{Matrix completion.}
Mathematically, a recommendation method can be posed as a {\em matrix completion} problem \cite{candes2012exact}, where columns and rows represent users and items, respectively, and matrix values represent a score determining whether a user would like an item or not. Given a small subset of known elements of the matrix, the goal is to fill in the rest. 
%
A famous example is the Netflix challenge \cite{art:KorenBellVolinsky09MatFac} offered in 2009 and carrying a 1M\$ prize for the algorithm that can best predict user ratings for movies based on previous ratings. The size of the Netflix is 480k movies $\times$ 18k users (8.5B entries), with only 0.011\% known entries.

Recently, there have been several attempts to incorporate geometric structure into matrix completion problems \cite{art:MaZhouLiuLyuKing11RecomSys,kalofolias2014matrix,rao2015collaborative,kuang2016harmonic}, e.g. in the form of column and row graphs representing similarity of users and items, respectively. Such additional information makes well-defined e.g. the notion of {\em smoothness} of data and was shown beneficial for the performance of recommender systems. 
These approaches can be generally related to the field of {\em signal processing on graphs} \cite{shuman2013emerging}, extending classical harmonic analysis methods to non-Euclidean domains.

\paragraph*{Geometric deep learning.} 
Of key interest to the design of recommender systems are deep learning approaches. 
In the recent years, deep neural networks and, in particular, convolutional neural networks (CNNs) \cite{lecun1998gradient} have been applied with great success to numerous computer vision-related applications. However, original CNN models cannot be directly applied to the recommendation problem to extract meaningful patterns in users, items and ratings because these data are not Euclidean structured, i.e. they do not lie on regular lattices like images but irregular domains like graphs or manifolds. This strongly motivates the development of {\em geometric deep learning} \cite{review_new} techniques that can mathematically deal with graph-structured data, which arises in numerous applications, ranging from computer graphics and vision \cite{masci2015geodesic,WFT2015,add16,boscaini2016learning,monti2016geometric} to chemistry \cite{duv2015convolutional}.

The earliest attempts to apply neural networks to graphs are due to Scarselli \etal \citeyear{gori2005new,GNN} (see more recent formulation \cite{GGSNN,comnets}). 
Bruna \etal \citeyear{bruna2013spectral,henaff2015deep} formulated CNN-like  deep neural architectures on graphs in the spectral domain, employing the analogy between the classical Fourier transforms and projections onto the eigenbasis of the graph Laplacian operator \cite{shuman2013emerging}. 
In a follow-up work, Defferrard \etal \citeyear{defferrard2016convolutional} proposed an efficient filtering scheme using recurrent Chebyshev polynomials, which reduces the complexity of CNNs on graphs to the same complexity of standard CNNs (on grids). This model was later extended to deal with dynamic data \cite{seo2016structured}. 
Kipf and Welling \citeyear{welling2016} proposed a simplification of Chebychev networks using simple filters operating on 1-hop neighborhoods of the graph.
Monti \etal \citeyear{monti2016geometric} introduced a spatial-domain generalization of CNNs to graphs local patch operators represented as Gaussian mixture models, showing a significant advantage of such models in generalizing across different graphs.

\paragraph*{Main contribution.} In this work, we treat matrix completion problem as deep learning on graph-structured data. 
We introduce a novel neural network architecture that is able to extract local stationary patterns from the high-dimensional spaces of users and items, and use these meaningful representations to infer the non-linear temporal diffusion mechanism of ratings. 
The spatial patterns are extracted by a new CNN architecture designed to work on multiple graphs. The temporal dynamics of the rating diffusion is produced by a Long-Short Term Memory (LSTM) recurrent neural network (RNN) \cite{art:HochreiterSchmidhuber97LSTM}. 
To our knowledge, our work is the first application of graph-based deep learning to matrix completion problem.

The rest of the paper is organized as follows. 
Section \ref{sec:rev} reviews the matrix completion models. Section \ref{sec:our_model} presents the proposed approach. Section \ref{sec:results} presents experimental results demonstrating the efficiency of our techniques on synthetic and real-world datasets, and Section~\ref{sec:conc} concludes the paper. 

\section{Background}
\label{sec:rev}

\subsection{Matrix Completion}
\paragraph*{Matrix completion problem.}
Recovering the missing values of a matrix given a small fraction of its entries is an ill-posed problem without additional mathematical constraints on the space of solutions. A well-posed problem is to assume that the variables lie in a smaller subspace, i.e., that the matrix is of low rank, 
\begin{eqnarray}
\min_{\mathbf{X}} \ \textrm{rank}(\mathbf{X}) \quad \textrm{s.t.} \quad x_{ij}=y_{ij},\  \forall ij \in \Omega,
\label{eq:lowrank}
\end{eqnarray}
 where $\mathbf{X}$ denotes the matrix to recover, $\Omega$ is the set of the known entries and $y_{ij}$ are their values. To make   \eqref{eq:lowrank} robust against noise and perturbation, the equality constraint can be replaced with a penalty 
  \begin{eqnarray}
 \min_{\mathbf{X}} \ \textrm{rank}(\mathbf{X}) + \frac{\mu}{2} \| \boldsymbol{\Omega} \circ (\mathbf{X}-\mathbf{Y}) \|_\mathrm{F}^2, 
  \end{eqnarray}
 where $\boldsymbol{\Omega}$ is the indicator matrix of the known entries $\Omega$ and $\circ$ denotes the Hadamard pointwise product.

Unfortunately, rank minimization turns out an NP-hard  combinatorial problem  that is computationally intractable in practical cases. The tightest possible convex relaxation of the previous problem is  
  \begin{eqnarray}
 \min_{\mathbf{X}}  \ \| \mathbf{X} \|_\star + \frac{\mu}{2} \|\boldsymbol{\Omega} \circ (\mathbf{X}-\mathbf{Y}) \|_\mathrm{F}^2,
  \end{eqnarray}
where $\| \cdot \|_\star$ is the nuclear norm of a matrix equal to the sum of its singular values \cite{art:CandesRecht09MatrixComple}. 
Cand{\`e}s and Recht \citeyear{art:CandesRecht09MatrixComple} proved that the $\ell_1$ relaxation of the SVD lead to solutions that recover almost exactly the original low-rank matrix.

\paragraph*{Geometric matrix completion}
An alternative relaxation of the rank operator in \eqref{eq:lowrank} is to constraint the space of solutions to be smooth w.r.t. some geometric structure of the matrix rows and columns \cite{art:MaZhouLiuLyuKing11RecomSys,kalofolias2014matrix,rao2015collaborative,art:BenziKalofoliasBressonVandergheynst16NMFTV}. 
The simplest model is proximity structure represented as an undirected weighted column graph 
 $\mathcal{G}_c = (\{1,\hdots, n\}, \mathcal{E}_c, \mathbf{W}_c)$ with {\em adjacency matrix} $\mathbf{W}_c = (w^\mathrm{c}_{ij})$, where $w^\mathrm{c}_{ij} = w^\mathrm{c}_{ji}$, $w^\mathrm{c}_{ij} = 0$ if $(i,j) \notin \mathcal{E}_c$ and $w^\mathrm{c}_{ij} > 0$ if $(i,j) \in \mathcal{E}_c$. 
 In our notation, the column graph could be thought of as a social network capturing relations between users and the similarity of their tastes. 
The row graph  $\mathcal{G}_r = (\{1,\hdots, m\}, \mathcal{E}_r, \mathbf{W}_r )$ representing the similarities of the items is defined in a similar manner.

On each of these graphs one can construct the (unnormalized) {\em graph Laplacian}, an $n\times n$ symmetric positive-semidefinite matrix $\boldsymbol{\Delta} = \mathbf{I} - \mathbf{D}^{-1/2} \mathbf{W} \mathbf{D}^{-1/2}$, where $\mathbf{D} = \mathrm{diag}\left(\sum_{j\neq i} w_{ij} \right)$ is the {\em degree matrix}. We denote the Laplacian associated with row and column graphs by $\boldsymbol{\Delta}_r$ and $\boldsymbol{\Delta}_c$, respectively. 
Considering the columns (respectively, rows) of matrix $\mathbf{X}$ as vector-valued functions on the column graph $\mathcal{G}_c$ (respectively, row graph $\mathcal{G}_r$), their smoothness can be expressed as the {\em Dirichlet norm} 
$\| \mathbf{X} \|_{\mathcal{G}_{r}}^2=\mathrm{trace}(\mathbf{X}^\top \boldsymbol{\Delta}_r \mathbf{X})$ (respecitvely, $\| \mathbf{X} \|_{\mathcal{G}_{c}}^2=\mathrm{trace}(\mathbf{X} \boldsymbol{\Delta}_c \mathbf{X}^\top)$). 
%
%
The {\em geometric matrix completion} problem thus boils down to minimizing 
 \begin{eqnarray}
\min_\mathbf{X} \ \| \mathbf{X} \|_{\mathcal{G}_r}^2 + \| \mathbf{X} \|_{\mathcal{G}_c}^2 + \frac{\mu}{2} \| \boldsymbol{\Omega} \circ (\mathbf{X} - \mathbf{Y}) \|_\mathrm{F}^2,
\label{eq:Dir}
\end{eqnarray}

%
%
%
%

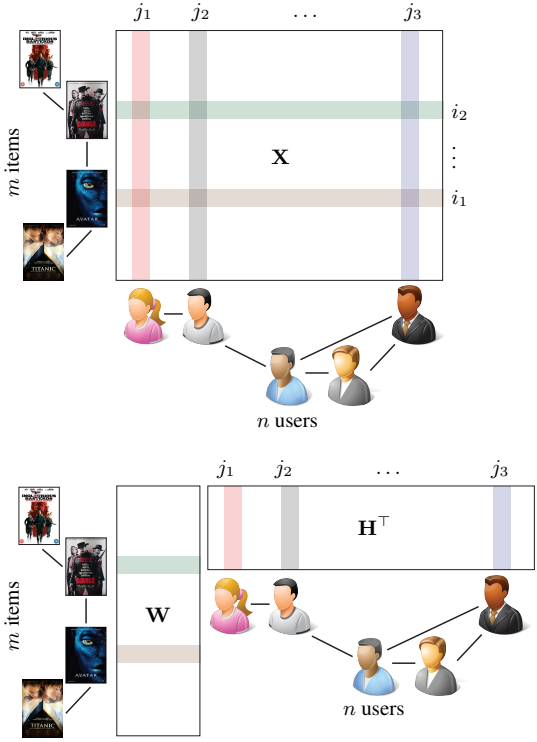
\begin{figure}[h!]
\centering
\scalebox{.88}{		
\begin{overpic}
	[trim=0cm 0cm 0cm 0cm,clip,width=1\linewidth]{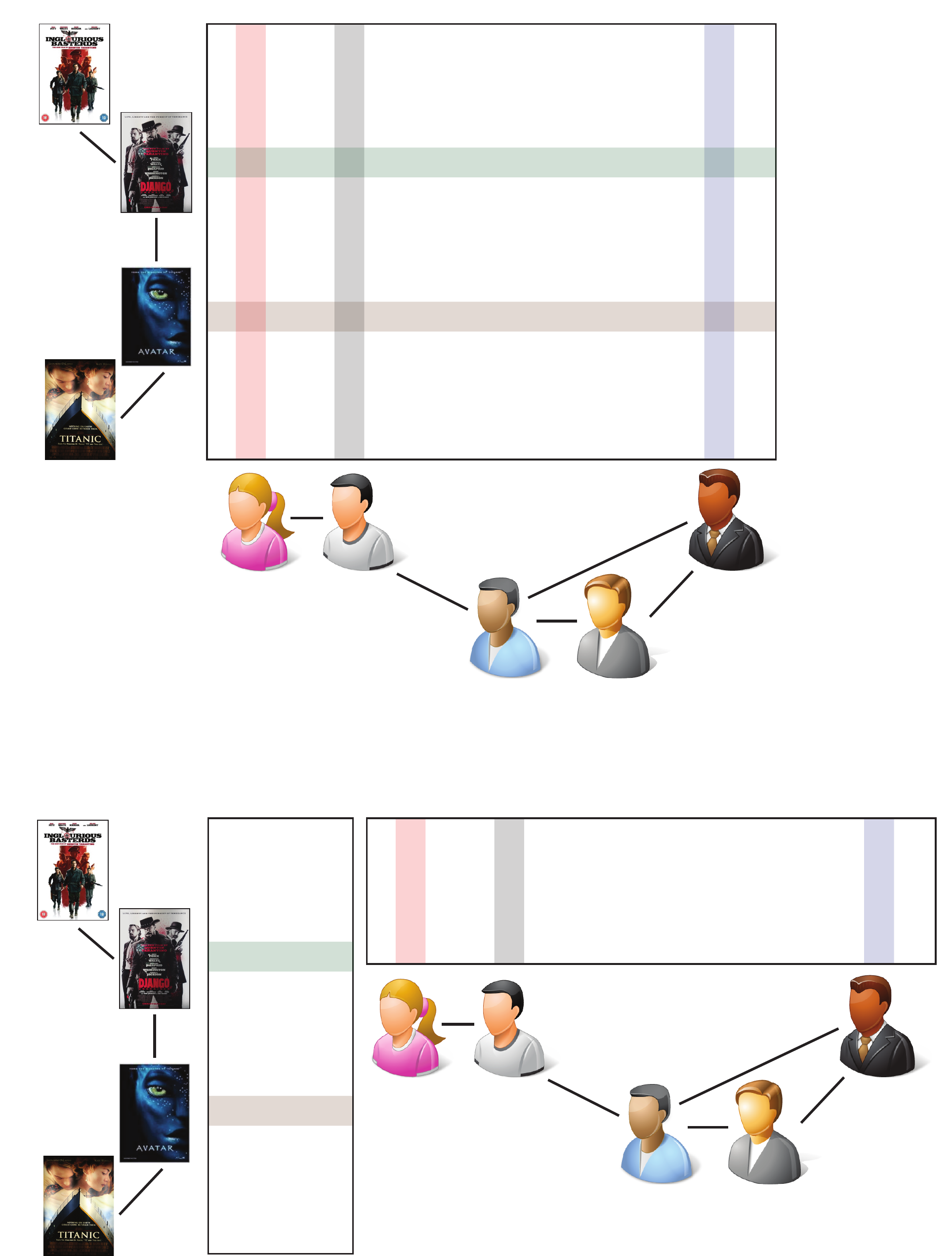}
			\put(36,43){\footnotesize $n$ users}
			\put(1,75){\rotatebox{90}{\footnotesize $m$ items}}
			\put(38,79.5){\footnotesize $\mathbf{X}$}			
			\put(20.5,16){\footnotesize $\mathbf{W}$}	
			\put(50,28){\footnotesize $\mathbf{H}^\top$}
			\put(19,100){\footnotesize $j_1$}
			\put(27,100){\footnotesize $j_2$}
			\put(41,100){\footnotesize $\hdots$}
			\put(56.5,100){\footnotesize $j_3$}
			\put(63,86){\footnotesize $i_2$}
			\put(63,79){\footnotesize $\vdots$}
			\put(63,74){\footnotesize $i_1$}
			\put(48,3){\footnotesize $n$ users}
			\put(1,12){\rotatebox{90}{\footnotesize $m$ items}}
			\put(30.5,36.5){\footnotesize $j_1$}
			\put(38.5,36.5){\footnotesize $j_2$}
			\put(52.5,36.5){\footnotesize $\hdots$}
			\put(68.5,36.5){\footnotesize $j_3$}
	\end{overpic}
	}
\caption{Full (top) and factorized (bottom) geometric matrix completion representations. 
The column and row graphs represent the relationships between users and items, respectively.  
 }
\label{fig:matrices}
\end{figure}

\paragraph*{Factorized models. }
Matrix completion algorithms introduced in the previous section are well-posed as convex optimization problems, guaranteeing existence, uniqueness and robustness of solutions. Besides, fast algorithms have been developed in the context of compressed sensing to solve the non-differential nuclear norm problem. However, the variables in this formulation are the full  $m\times n$ matrix $\mathbf{X}$, making such methods hard to scale up to large matrices such as the notorious Netflix challenge. 

A solution is to use a factorized representation \cite{art:SrebroRennieJaakkola04MatFact,art:KorenBellVolinsky09MatFac,art:MaZhouLiuLyuKing11RecomSys,art:YanezBach14NMF,rao2015collaborative,art:BenziKalofoliasBressonVandergheynst16NMFTV} 
$\mathbf{X}=\mathbf{W}\mathbf{H}^\top$, 
where $\mathbf{W}, \mathbf{H}$ are $m\times r$ and $n\times r$ matrices, respectively, with $r\ll \min(m,n)$. The use of factors $\mathbf{W}, \mathbf{H}$ reduce the number of degrees of freedom from $\mathcal{O}(mn)$ to $\mathcal{O}(m+n)$; this representation is also attractive as solving the matrix completion problem often assumes the original matrix to be low-rank, and $\mathrm{rank}(\mathbf{W}\mathbf{H}^\top)\leq r$ by construction. 
Figure \ref{fig:matrices} shows the full and factorized settings of the matrix completion problem.

The nuclear norm minimization problem in the previous section can be equivalently rewritten in a factorized form as \cite{art:SrebroRennieJaakkola04MatFact}:
 \begin{eqnarray}
\min_{\mathbf{W},\mathbf{H}} \ \frac{1}{2} \| \mathbf{W} \|_\mathrm{F}^2 + \frac{1}{2} \| \mathbf{H} \|_\mathrm{F}^2 + \frac{\mu}{2} \| \boldsymbol{\Omega} \circ (\mathbf{W}\mathbf{H}^\top - \mathbf{Y}) \|_\mathrm{F}^2.
\label{eq:candes_fact}
\end{eqnarray}
and the factorized formulation of the graph-based minimization problem \eqref{eq:Dir} as 
 \begin{eqnarray}
\min_{\mathbf{W},\mathbf{H}}  \ \frac{1}{2} \| \mathbf{W} \|_{\mathcal{G}_{r}}^2 + \frac{1}{2} \| \mathbf{H} \|_{\mathcal{G}_{c}}^2 + \frac{\mu}{2} \| \boldsymbol{\Omega} \circ (\mathbf{W}\mathbf{H}^\top - \mathbf{Y}) \|_\mathrm{F}^2.
\label{eq:Dir_fact}
\end{eqnarray}
The limitation of model \eqref{eq:Dir_fact} is to decouple the regularization process applied simultaneously on the rows and columns of $\mathbf{X}$ in \eqref{eq:Dir}, but the advantage is linear instead of quadratic complexity. 

\subsection{Deep learning on graphs}
The key idea to our work is {\em geometric deep learning}, an extension of the popular CNNs to graphs. 
A graph Laplacian admits a spectral eigendecomposition of the form 
$
\boldsymbol{\Delta} = \boldsymbol{\Phi} \boldsymbol{\Lambda} \boldsymbol{\Phi}^\top
$, 
where $\boldsymbol{\Phi} = (\boldsymbol{\phi}_1, \hdots \boldsymbol{\phi}_n)$ denotes the matrix of orthonormal eigenvectors and $\boldsymbol{\Lambda} = \mathrm{diag}(\lambda_1, \hdots, \lambda_n)$ is the diagonal matrix of the corresponding eigenvalues. The eigenvectors play the role of Fourier atoms in classical harmonic analysis and the eigenvalues can be interpreted as frequencies. 
%
Given a function $\mathbf{x} = (x_1, \hdots, x_n)^\top$ on the vertices of the graph, its {\em graph Fourier transform} is given by $\hat{\mathbf{x}} = \boldsymbol{\Phi}^\top\mathbf{x}$. 
The {\em spectral convolution} of two functions $\mathbf{x}, \mathbf{y}$ can be defined as the element-wise product of the respective  Fourier transforms,
\begin{equation} 
\label{spectral_conv}
\mathbf{x} \star \mathbf{y} = \boldsymbol{\Phi}(\boldsymbol{\Phi}^\top\mathbf{x}) \circ (\boldsymbol{\Phi}^\top\mathbf{y}) = \boldsymbol{\Phi}\, \mathrm{diag}(\hat{y}_1, \hdots, \hat{y}_n)\,\hat{\mathbf{x}}. 
\end{equation}
%


Bruna \etal \citeyear{bruna2013spectral} used the spectral definition of convolution~(\ref{spectral_conv}) to generalize CNNs on graphs. 
A spectral convolutional layer has the form 
\begin{equation} 
\label{spectral_construction_eq}
\tilde{\mathbf{x}}_l =   \xi \left(  \sum_{l'=1}^{q'} \boldsymbol{\Phi} \hat{\mathbf{Y}}_{ll'} \boldsymbol{\Phi}^\top \mathbf{x}_{l'} \right), \hspace{3mm} l = 1,\hdots, q,
\end{equation}
where $q', q$ denote the number of input and output channels, respectively,  
%
%
$\hat{\mathbf{Y}}_{ll'} = \mathrm{diag}(\hat{y}_{ll',1}, \hdots, \hat{y}_{ll',n})$ is a diagonal matrix of spectral multipliers representing a learnable filter in the spectral domain, and $\xi$ is a nonlinearity (e.g. ReLU) applied on the vertex-wise function values. 
%
%
%
%
%
%
Unlike classical convolutions carried out efficiently in the spectral domain using FFT, the computations of the forward and inverse graph Fourier transform incur expensive $\mathcal{O}(n^2)$ multiplication by the matrices $\boldsymbol{\Phi}, \boldsymbol{\Phi}^\top$, as there are no FFT-like algorithms on general graphs. 
Furthermore, there is no guarantee that the filters represented in the spectral domain are localized in the spatial domain, which is an important property of classical CNNs.  

\begin{figure*}[ht!]
\centering
\vspace{1mm}
\scalebox{.8}{
\vspace{5mm}
		\begin{overpic}
	[trim=0cm 0cm 0cm 0cm,clip,width=1\linewidth]{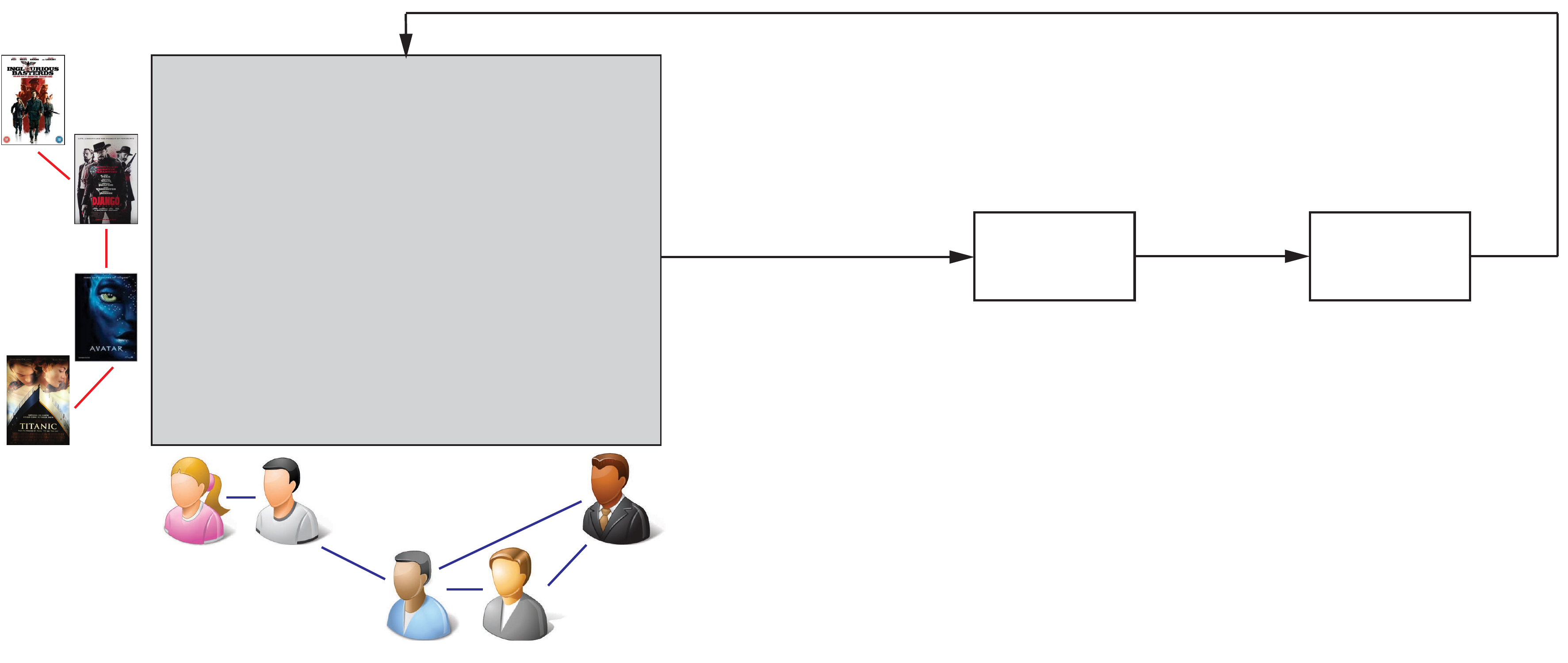}
			\put(25,26){\footnotesize $\mathbf{X}$}	
			\put(55,27){\footnotesize $\mathbf{X}^{(t)}$}	
			\put(76,27){\footnotesize $\tilde{\mathbf{X}}^{(t)}$}	
			\put(64,25.25){\footnotesize MGCNN}	
			\put(86.5,25.25){\footnotesize RNN}	
			\put(100,33.25){\footnotesize $\mathbf{dX}^{(t)}$}	
			\put(68,42.5){\footnotesize $\mathbf{X}^{(t+1)} = \mathbf{X}^{(t)} + \mathbf{dX}^{(t)}$}	
			\put(70,20){\footnotesize \color{gray}{row+column filtering}}
	\end{overpic}\vspace{-2mm}
}
\caption{Recurrent GCNN (RGCNN) architecture using the full matrix completion model and operating simultaneously on the rows and columns of the matrix $\mathbf{X}$. The output of the Multi-Graph CNN (MGCNN) module is a $q$-dimensional feature vector for each element of the input matrix. The number of parameters to learn is $\mathcal{O}(1)$ and the learning complexity is $\mathcal{O}(mn)$.
 }
\label{fig:architectureX}
\end{figure*}
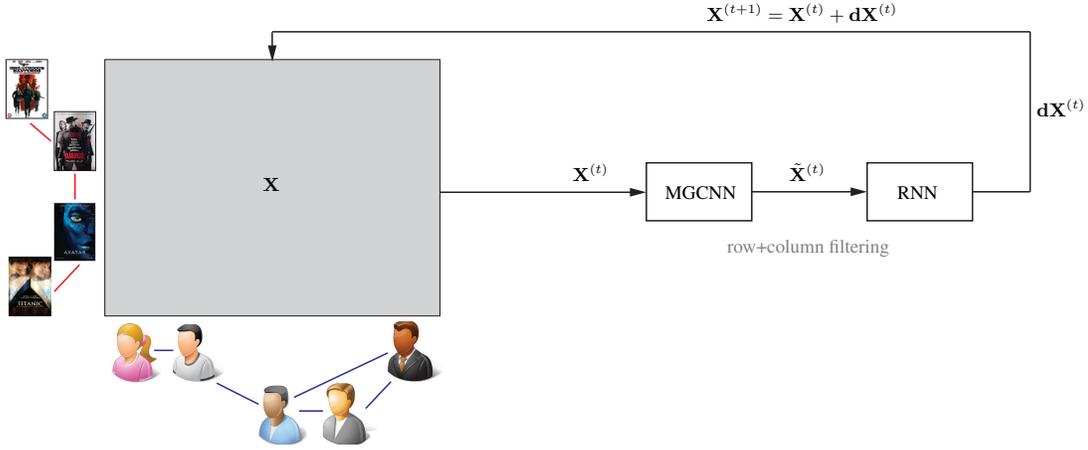

To address these issues, Defferrard \etal \citeyear{defferrard2016convolutional} proposed using an explicit expansion in the Chebyshev polynomial basis to represent the spectral filters\vspace{-1mm}
\begin{equation} \label{eq:filt_cheby}
	\tau_{\boldsymbol{\theta}}(\tilde{\boldsymbol{\Delta}}) = \sum_{j=0}^{p-1} \theta_j T_j(\tilde{\boldsymbol{\Delta}}) = \sum_{j=0}^{p-1} \theta_j  \boldsymbol{\Phi} T_j(\tilde{\boldsymbol{\Lambda}})\boldsymbol{\Phi}^\top,
\end{equation}
where $\tilde{\boldsymbol{\Delta}} = 2 \lambda_{n}^{-1}\boldsymbol{\Delta}  - \mathbf{I}$ is the rescaled Laplacian such that its eigenvalues $\tilde{\boldsymbol{\Lambda}} = 2 \lambda_{n}^{-1} \boldsymbol{\Lambda}  - \mathbf{I}$ are in the interval $[-1,1]$, 
$\boldsymbol{\theta}$ is the $p$-dimensional vector of polynomial coefficients parametrizing the filter, and $T_j(\lambda) = 2\lambda T_{j-1}(\lambda) - T_{j-2}(\lambda)$ denotes the Chebyshev polynomial of degree $j$ defined in a recursive manner with $T_1(\lambda) =\lambda$ and $T_0(\lambda) =1$. \footnote{$T_j(\lambda) = \cos(j \cos^{-1}(\lambda))$ is an oscillating function on $[-1, 1]$ with $j$ roots, $j+1$ equally spaced extrema, and a frequency linearly dependent on $j$. Chebyshev polynomials form an orthogonal basis for the space of smooth functions on $[-1,1]$ and are thus convenient to compactly represent spectral filters.}
This approach benefits from several advantages. First, it does not require an explicit computation of the Laplacian eigenvectors, and due to the recursive definition of the Chebyshev polynomials, the computation of the filter 
incurs applying the Laplacian $p$ times. Multiplication by Laplacian has the cost of $\mathcal{O}(|\mathcal{E}|)$, and assuming the graph has $|\mathcal{E}| = \mathcal{O}(n)$ edges (which is the case for $k$-nearest neighbors graphs and most real-world networks), the overall complexity is $\mathcal{O}(n)$ rather than $\mathcal{O}(n^2)$ operations, which is the same complexity than standard CNNs.
Moreover, since the Laplacian is a local operator affecting only 1-hop neighbors of a vertex and accordingly its $(p-1)$st power affects the $p$-hop neighborhood, the resulting filters are spatially localized.

\section{Our approach}
\label{sec:our_model}

\begin{figure*}[ht!]
\centering
\vspace{1mm}
\scalebox{.8}{
\vspace{5mm}
		\begin{overpic}
	[trim=0cm 0cm 0cm 0cm,clip,width=1\linewidth]{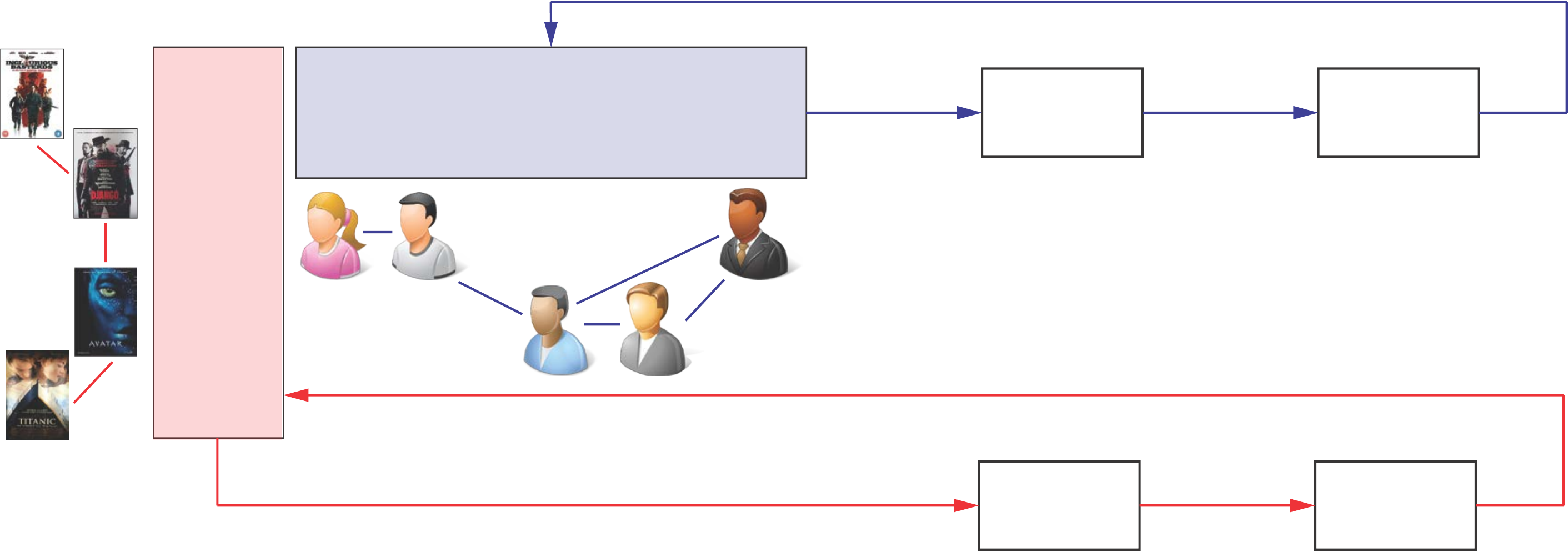}
			\put(13.5,19){\footnotesize $\mathbf{W}$}	
			\put(34,27.5){\footnotesize $\mathbf{H}^\top$}	
			\put(55,29.5){\footnotesize $\mathbf{H}^{(t)}$}	
			\put(76,29.5){\footnotesize $\tilde{\mathbf{H}}^{(t)}$}	
			\put(55,4.5){\footnotesize $\mathbf{W}^{(t)}$}	
			\put(76,4.5){\footnotesize $\tilde{\mathbf{W}}^{(t)}$}	
			\put(64.75,27.75){\footnotesize GCNN}	
			\put(86.5,27.75){\footnotesize RNN}	
			\put(64.75,2.75){\footnotesize GCNN}	
			\put(86.5,2.75){\footnotesize RNN}	
			\put(100,31.25){\footnotesize $\mathbf{dH}^{(t)}$}	
			\put(100,6.25){\footnotesize $\mathbf{dW}^{(t)}$}
			\put(68,11.75){\footnotesize $\mathbf{W}^{(t+1)} = \mathbf{W}^{(t)} + \mathbf{dW}^{(t)}$}	
			\put(68,36.5){\footnotesize $\mathbf{H}^{(t+1)} = \mathbf{H}^{(t)} + \mathbf{dH}^{(t)}$}	
			\put(73,-1){\footnotesize {\color{red}{row filtering}}}	
			\put(72,23){\footnotesize {\color{blue}{column filtering}}}	
	\end{overpic}
	}
\caption{Separable Recurrent GCNN (sRGCNN) architecture using the factorized matrix completion model and operating separately on the rows and columns of the factors $\mathbf{W}$, $\mathbf{H}^\top$. The output of the GCNN module is a $q$-dimensional feature vector for each input row/column, respectively. The number of parameters to learn is $\mathcal{O}(1)$ and the learning complexity is $\mathcal{O}(m+n)$.
 }
\label{fig:architectureWH}
\end{figure*}
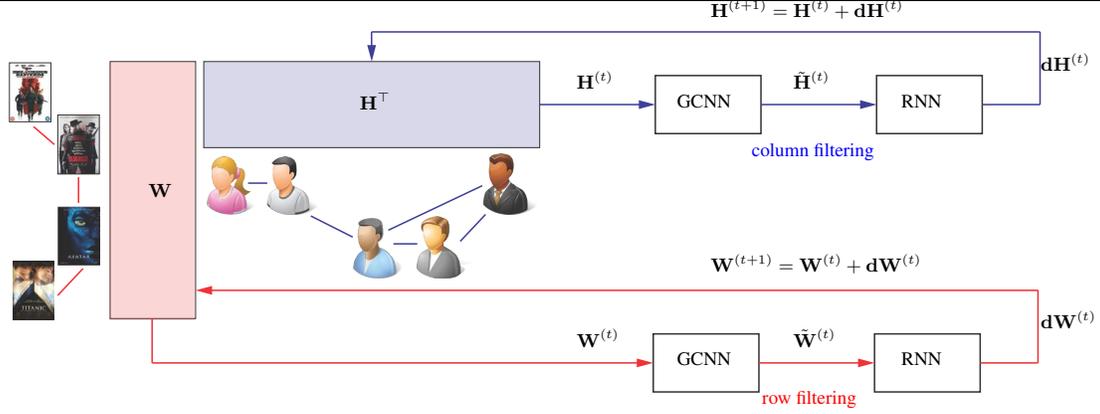

In this paper, we propose formulating matrix completion as a learnable diffusion process applied to the score values. 
The deep learning architecture considered for this purpose consists of a spatial part extracting spatial features from the matrix (we consider two different approaches working on the full and factorized matrix models), and a temporal part using a recurrent LSTM network. The two architectures are summarized in Figures~\ref{fig:architectureX} and~\ref{fig:architectureWH} and described in details in the following.

\subsection{Multi-Graph CNNs}


\paragraph*{Multi-graph convolution. }
Our first goal is to extend the notion of the aforementioned graph Fourier transform to matrices whose rows and columns are defined on row- and column-graphs. We recall that a classical two-dimensional Fourier transform of an image (matrix) can be thought of as applying a one-dimensional Fourier transform to its rows and columns. 
In our setting, the analogy of the two-dimensional Fourier transform has the form
\begin{equation}
\hat{\mathbf{X}} = \boldsymbol{\Phi}_r^\top \mathbf{X} \boldsymbol{\Phi}_c
\end{equation}
where $\boldsymbol{\Phi}_c, \boldsymbol{\Phi}_r$ and $\boldsymbol{\Lambda}_c, \boldsymbol{\Lambda}_r$  denote the $n\times n$ and $m\times m$ eigenvector- and eigenvalue matrices of the column- and row-graph Laplacians $\boldsymbol{\Delta}_c, \boldsymbol{\Delta}_r$, respectively. 
The multi-graph version of the spectral convolution~(\ref{spectral_conv}) is given by 
\begin{equation}
\mathbf{X} \star \mathbf{Y} =   \boldsymbol{\Phi}_r (\hat{\mathbf{X}} \circ \hat{\mathbf{Y}}) \boldsymbol{\Phi}_c^\top. 
\end{equation}
Representing the filters as their spectral multipliers $\hat{\mathbf{Y}}$ would yield $\mathcal{O}(mn)$ parameters, prohibitive in any practical application. 
To overcome this limitation, we resort to the representation of the filters in Chebychev polynomial bases of degree $p$, 
\begin{equation}
	\tau_{\boldsymbol{\Theta}}(\tilde{\lambda}_c, \tilde{\lambda}_r) = \sum_{j,j'=0}^{p} \theta_{jj'} T_j(\tilde{\lambda}_c) T_{j'}(\tilde{\lambda}_r),
\end{equation}
where $\boldsymbol{\Theta} = (\theta_{jj'})$ is the $(p+1)\times(p+1)$ matrix of coefficients, i.e., $\mathcal{O}(1)$ parameters.  
The application of such filters to the matrix $\mathbf{X}$ 
\begin{equation}
\tilde{\mathbf{X}} = \sum_{j,j' = 0}^p \theta_{jj'} T_{j}(\tilde{\boldsymbol{\Delta}}_r) \mathbf{X} T_{j'}(\tilde{\boldsymbol{\Delta}}_c) 
\label{eq:cheb-approx}
\end{equation}
results in $\mathcal{O}(mn)$ complexity. Here, as previously, $\tilde{\boldsymbol{\Delta}}_r$, $\tilde{\boldsymbol{\Delta}}_c$ denote the scaled Laplacians with eigenvalues in the interval $[-1,1]$. 


A {\em Multi-Graph CNN} (MGCNN) using this parametrization of filters~(\ref{eq:cheb-approx}) in the convolutional layer is applied to the $m\times n$ matrix $\mathbf{X}$ (single input channel), producing $q$ outputs (i.e., a tensor of size $m\times n \times q$).

\paragraph*{Separable convolution. } 
A simplification of the multi-graph convolution is obtained considering the factorized form of the matrix $\mathbf{X} = \mathbf{W}\mathbf{H}^\top$ and applying one-dimensional convolution on the respective graph to each factor, 
%
\begin{eqnarray}
\tilde{\mathbf{w}}_{l} &=& \sum_{l'=1}^{q'} \boldsymbol{\Phi}_r \hat{\mathbf{Y}}_r^{ll'} \boldsymbol{\Phi}_r^\top{\mathbf{w}}_{l'}, \hspace{3mm} l = 1,\hdots, q\\
\tilde{\mathbf{h}}_{l} &=& \sum_{l'=1}^{q'} \boldsymbol{\Phi}_c \hat{\mathbf{Y}}_c^{ll'} \boldsymbol{\Phi}_c^\top{\mathbf{h}}_{l'}, \hspace{3mm} l = 1,\hdots, q,
\end{eqnarray}

%
where $\hat{\mathbf{Y}}_r^{ll'} = \mathrm{diag}(\hat{y}^r_{ll',1}, \hdots, \hat{y}^r_{ll',m})$ and $\hat{\mathbf{Y}}_c^{ll'} = \mathrm{diag}(\hat{y}^c_{ll',1}, \hdots, \hat{y}^c_{ll',n})$ are the row- and column-filters resulting in a total of $\mathcal{O}(m+n)$ parameters. 
Similarly to the previous case, we can express the filters resorting to Chebyshev polynomials, 
\begin{eqnarray}
\tilde{\mathbf{w}}_{l} &=& \sum_{l'=1}^{q'} \sum_{j=0}^p \theta_{ll', j}^r T_{j}(\tilde{\boldsymbol{\Delta}}_r) \mathbf{w}_{l'}\\
\tilde{\mathbf{h}}_{l} &=& \sum_{l'=1}^{q'} \sum_{j'=0}^p \theta_{ll', j'}^c T_{j'}(\tilde{\boldsymbol{\Delta}}_c) \mathbf{h}_{l'}
\label{eq:cheb-approxs}
\end{eqnarray}
with $2(p+1)qq'$ parameters in total.

\subsection{Matrix diffusion with RNN}

%
%
%
%
%

The next step of our approach is to feed the features extracted from the matrix by the MGCNN (or alternatively, the row- and column-GCNNs) to a Recurrent Neural Network (RNN) implementing the score diffusion process. We use the classical Long-Short Term Memory (LSTM) RNN architecture \cite{art:HochreiterSchmidhuber97LSTM}, which has demonstrated to be highly efficient to learn the dynamical property of data sequences as LSTM is able to keep long-term internal states (in particular, avoiding the vanishing gradient issue). The input of the LSTM gate is given by the static features extracted from the MGCNN, which can be seen as a projection or dimensionality reduction of the original matrix in the space of the most meaningful and representative information (the disentanglement effect). This representation coupled with LSTM appears particularly well-suited to keep a long term internal state, which allows to predict accurate small changes $\mathbf{dX}$ of the matrix $\mathbf{X}$ (or $\mathbf{dW}$, $\mathbf{dH}$ of the factors $\mathbf{W}$, $\mathbf{H}$) that can propagate through the full temporal steps. 

Figures~\ref{fig:architectureX} and~\ref{fig:architectureWH} provides an illustration of the proposed matrix completion model. We also give a precise description of the two settings of our model in Algorithms~\ref{alg:RGCNN} and~\ref{alg:sRGCNN}. 
We refer to the whole architecture combining the MGCNN and RNN in the full matrix completion setting as Recurrent Graph CNN (RGCNN). The factorized version with two GCNNs and RNN is referred to as separable Recurrent Graph CNN (sRGCNN). 

\begin{algorithm}[!t]
\caption{Full matrix completion model using RGCNN}\label{alg:RGCNN}
\begin{algorithmic}[1]
\INPUT $m\times n$ matrix $\mathbf{X}^{(0)}$ containing initial values 
\FOR{$t = 0 : T$}
	\STATE Apply the Multi-Graph CNN \eqref{eq:cheb-approx} on $\mathbf{X}^{(t)}$ producing an $m\times n \times q$ output $\tilde{\mathbf{X}}^{(t)}$ containing a $q$-dimensional feature vector for each matrix element.
	\FOR {all elements $(i,j)$}
		\STATE Apply RNN to feature vector $\tilde{\mathbf{x}}^{(t)}_{ij} = (\tilde{x}^{(t)}_{ij1}, \hdots, \tilde{x}^{(t)}_{ijq})$ producing the predicted incremental value $dx^{(t)}_{ij}$
	\ENDFOR
	\STATE Update $\mathbf{X}^{(t+1)} = \mathbf{X}^{(t)} + \mathbf{dX}^{(t)}$
\ENDFOR
\end{algorithmic}
\end{algorithm}

\begin{algorithm}[!t]
\caption{Factorized matrix completion model using sRGCNN}\label{alg:sRGCNN}
\begin{algorithmic}[1]
\INPUT $m\times r$ factor $\mathbf{H}^{(0)}$ and $n\times r$ factor $\mathbf{W}^{(0)}$ representing the matrix $\mathbf{X}^{(0)}$
\FOR{$t = 0 : T$}
	\STATE Apply the Graph CNN on $\mathbf{H}^{(t)}$ producing an $n \times q$ output $\tilde{\mathbf{H}}^{(t)}$.
	\FOR {$j=1:n$}
		\STATE Apply RNN to feature vector $\tilde{\mathbf{h}}^{(t)}_j = (\tilde{h}^{(t)}_{j1}, \hdots, \tilde{h}^{(t)}_{jq})$ producing the predicted incremental value $dh^{(t)}_{j}$
	\ENDFOR
	\STATE Update $\mathbf{H}^{(t+1)} = \mathbf{H}^{(t)} + \mathbf{dH}^{(t)}$
		\STATE Apply the Graph CNN on $\mathbf{W}^{(t)}$ producing an $m \times q$ output $\tilde{\mathbf{W}}^{(t)}$.
	\FOR {$i=1:m$}
		\STATE Apply RNN to feature vector $\tilde{\mathbf{w}}^{(t)}_i = (\tilde{w}^{(t)}_{i1}, \hdots, \tilde{w}^{(t)}_{iq})$ producing the predicted incremental value $dw^{(t)}_{i}$
	\ENDFOR
	\STATE Update $\mathbf{W}^{(t+1)} = \mathbf{W}^{(t)} + \mathbf{dW}^{(t)}$
\ENDFOR
\end{algorithmic}
\end{algorithm}

The complexity of Algorithm \ref{alg:RGCNN} scales quadratically as $\mathcal{O}(mn)$ due to the use of MGCNN. For large matrices, we can opt for Algorithm \ref{alg:sRGCNN} that processes the rows and columns separately with standard GCNNs and scales linearly as $\mathcal{O}(m+n)$. 

\subsection{Training}
Training of the networks is performed by minimizing the loss 
\begin{equation}
\ell(\boldsymbol{\Theta},\boldsymbol{\sigma}) = 
\| \mathbf{X}^{(T)}_{\boldsymbol{\Theta},\boldsymbol{\sigma}} \|_{\mathcal{G}_r}^2 + \| \mathbf{X}^{(T)}_{\boldsymbol{\Theta},\boldsymbol{\sigma}} \|_{\mathcal{G}_c}^2 + \frac{\mu}{2} \| \boldsymbol{\Omega}\circ (\mathbf{X}^{(T)}_{\boldsymbol{\Theta},\boldsymbol{\sigma}} - \mathbf{Y}) \|_\mathrm{F}^2. 
\end{equation}
Here, $T$ denotes the number of diffusion iterations (applications of the RNN), and we use the notation $\mathbf{X}^{(T)}_{\boldsymbol{\Theta},\boldsymbol{\sigma}}$ to emphasize that the matrix depends on the parameters of the MGCNN (Chebyshev polynomial coefficients $\boldsymbol{\Theta}$) and those of the LSTM (denoted by $\boldsymbol{\sigma}$). 

In the factorized setting, we use the loss 
\begin{eqnarray}
\ell(\boldsymbol{\theta}_r, \boldsymbol{\theta}_c,\boldsymbol{\sigma}) &=& 
\| \mathbf{W}^{(T)}_{\boldsymbol{\theta}_r, \boldsymbol{\sigma}} \|_{\mathcal{G}_r}^2 + \| \mathbf{H}^{(T)}_{\boldsymbol{\theta}_c,\boldsymbol{\sigma}} \|_{\mathcal{G}_c}^2  \\
&+& \frac{\mu}{2} \| \boldsymbol{\Omega}\circ (\mathbf{W}^{(T)}_{\boldsymbol{\theta}_r,\boldsymbol{\sigma}} (\mathbf{H}^{(T)}_{\boldsymbol{\theta}_c,\boldsymbol{\sigma}} )^\top - \mathbf{Y}) \|_\mathrm{F}^2. \nonumber
\end{eqnarray}
%
where $\boldsymbol{\theta}_c, \boldsymbol{\theta}_r$ are the parameters of the two GCNNs. 

\section{Results}
\label{sec:results}

\label{sec:results}

\paragraph*{Experimental settings.} 
We closely followed the experimental setup of \cite{rao2015collaborative}, using five standard datasets: Synthetic dataset from \cite{kalofolias2014matrix}, MovieLens \cite{miller2003movielens}, Flixster \cite{jamali2010matrix}, Douban \cite{art:MaZhouLiuLyuKing11RecomSys}, and YahooMusic \cite{dror2012yahoo}. 
Classical Matrix Completion (MC) \cite{candes2012exact}, Inductive Matrix Completion (IMC) \cite{jain2013provable,xu2013speedup}, Geometric Matrix Completion (GMC) \cite{kalofolias2014matrix}, and Graph Regularized Alternating Least Squares (GRALS) \cite{rao2015collaborative} were used as baseline methods.

In all the experiments, we used the following settings for our RGCNNs: Chebyshev polynomials of order $p=5$, outputting $k=32$-dimensional features, LSTM cells with $32$ features and $T=10$ diffusion steps. All the models were implemented in Google TensorFlow and trained using the Adam stochastic optimization algorithm \cite{KingmaB14} with learning rate $10^{-3}$. In factorized models, rank $r=15$ and $10$ was used for the synthetic and real datasets, respectively. 
For all methods, hyperparameters were chosen by cross-validation. 


\subsection{Synthetic data}
\begin{figure*}[!ht]
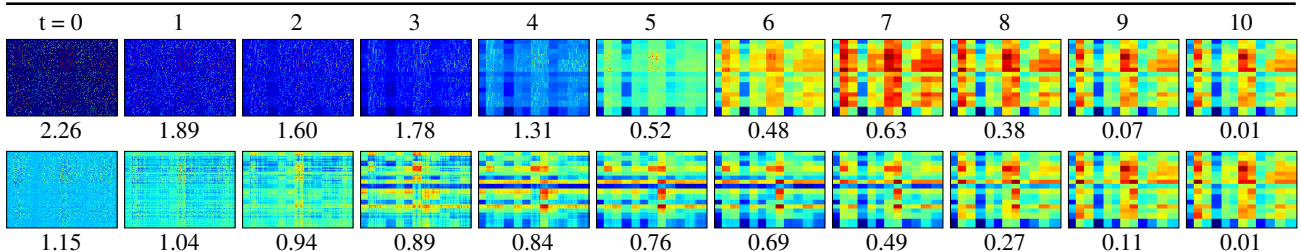

\centering
\setlength\figureheight{2.6cm}
\setlength\figurewidth{3.05cm}

\begin{minipage}[h]{1.0\linewidth}
\vspace{1mm}
\footnotesize
\vspace{-3mm}
\begin{minipage}[h]{0.085\linewidth}
\center
t = 0
\end{minipage}
\hfill
\begin{minipage}[h]{0.085\linewidth}
\center
1
\end{minipage}
\hfill
\begin{minipage}[h]{0.085\linewidth}
\center
2
\end{minipage}
\hfill
\begin{minipage}[h]{0.085\linewidth}
\center
3
\end{minipage}
\hfill
\begin{minipage}[h]{0.085\linewidth}
\center
4
\end{minipage}
\hfill
\begin{minipage}[h]{0.085\linewidth}
\center
5
\end{minipage}
\hfill
\begin{minipage}[h]{0.085\linewidth}
\center
6
\end{minipage}
\hfill
\begin{minipage}[h]{0.085\linewidth}
\center
7
\end{minipage}
\hfill
\begin{minipage}[h]{0.085\linewidth}
\center
8
\end{minipage}
\hfill
\begin{minipage}[h]{0.085\linewidth}
\center
9
\end{minipage}
\hfill
\begin{minipage}[h]{0.085\linewidth}
\center
10
\end{minipage}
\end{minipage}
\begin{tikzpicture}
\begin{groupplot}[
     group style = {group size = 11 by 1, horizontal sep=0.1cm}]
\input{Figures/final_results/tikz/synthetic_netflix/evolutions_X/time_0/X.tex}
\input{Figures/final_results/tikz/synthetic_netflix/evolutions_X/time_1/X.tex}
\input{Figures/final_results/tikz/synthetic_netflix/evolutions_X/time_2/X.tex}
\input{Figures/final_results/tikz/synthetic_netflix/evolutions_X/time_3/X.tex}
\input{Figures/final_results/tikz/synthetic_netflix/evolutions_X/time_4/X.tex}
\input{Figures/final_results/tikz/synthetic_netflix/evolutions_X/time_5/X.tex}
\input{Figures/final_results/tikz/synthetic_netflix/evolutions_X/time_6/X.tex}
\input{Figures/final_results/tikz/synthetic_netflix/evolutions_X/time_7/X.tex}
\input{Figures/final_results/tikz/synthetic_netflix/evolutions_X/time_8/X.tex}
\input{Figures/final_results/tikz/synthetic_netflix/evolutions_X/time_9/X.tex}
\input{Figures/final_results/tikz/synthetic_netflix/evolutions_X/time_10/X.tex}
  \end{groupplot}
\end{tikzpicture}
\begin{minipage}[h]{1.0\linewidth}
\footnotesize
\vspace{-3mm}
\begin{minipage}[h]{0.085\linewidth}
\center
2.26
\end{minipage}
\hfill
\begin{minipage}[h]{0.085\linewidth}
\center
 1.89
\end{minipage}
\hfill
\begin{minipage}[h]{0.085\linewidth}
\center
 1.60
\end{minipage}
\hfill
\begin{minipage}[h]{0.085\linewidth}
\center
 1.78
\end{minipage}
\hfill
\begin{minipage}[h]{0.085\linewidth}
\center
 1.31
\end{minipage}
\hfill
\begin{minipage}[h]{0.085\linewidth}
\center
 0.52
\end{minipage}
\hfill
\begin{minipage}[h]{0.085\linewidth}
\center
 0.48
\end{minipage}
\hfill
\begin{minipage}[h]{0.085\linewidth}
\center
 0.63
\end{minipage}
\hfill
\begin{minipage}[h]{0.085\linewidth}
\center
 0.38
\end{minipage}
\hfill
\begin{minipage}[h]{0.085\linewidth}
\center
 0.07
\end{minipage}
\hfill
\begin{minipage}[h]{0.085\linewidth}
\center
 0.01
\end{minipage}
\end{minipage}
\begin{tikzpicture}
\begin{groupplot}[
     group style = {group size = 11 by 1, horizontal sep=0.1cm}]
\input{Figures/final_results/tikz/synthetic_netflix/evolutions_X_2_diff_conv/time_0/X.tex}
\input{Figures/final_results/tikz/synthetic_netflix/evolutions_X_2_diff_conv/time_1/X.tex}
\input{Figures/final_results/tikz/synthetic_netflix/evolutions_X_2_diff_conv/time_2/X.tex}
\input{Figures/final_results/tikz/synthetic_netflix/evolutions_X_2_diff_conv/time_3/X.tex}
\input{Figures/final_results/tikz/synthetic_netflix/evolutions_X_2_diff_conv/time_4/X.tex}
\input{Figures/final_results/tikz/synthetic_netflix/evolutions_X_2_diff_conv/time_5/X.tex}
\input{Figures/final_results/tikz/synthetic_netflix/evolutions_X_2_diff_conv/time_6/X.tex}
\input{Figures/final_results/tikz/synthetic_netflix/evolutions_X_2_diff_conv/time_7/X.tex}
\input{Figures/final_results/tikz/synthetic_netflix/evolutions_X_2_diff_conv/time_8/X.tex}
\input{Figures/final_results/tikz/synthetic_netflix/evolutions_X_2_diff_conv/time_9/X.tex}
\input{Figures/final_results/tikz/synthetic_netflix/evolutions_X_2_diff_conv/time_10/X.tex}
  \end{groupplot}
\end{tikzpicture}
\begin{minipage}[h]{1.0\linewidth}
\footnotesize
\vspace{-3mm}
\begin{minipage}[h]{0.085\linewidth}
\center
 1.15
\end{minipage}
\hfill
\begin{minipage}[h]{0.085\linewidth}
\center
 1.04
\end{minipage}
\hfill
\begin{minipage}[h]{0.085\linewidth}
\center
 0.94
\end{minipage}
\hfill
\begin{minipage}[h]{0.085\linewidth}
\center
 0.89
\end{minipage}
\hfill
\begin{minipage}[h]{0.085\linewidth}
\center
 0.84
\end{minipage}
\hfill
\begin{minipage}[h]{0.085\linewidth}
\center
 0.76
\end{minipage}
\hfill
\begin{minipage}[h]{0.085\linewidth}
\center
 0.69
\end{minipage}
\hfill
\begin{minipage}[h]{0.085\linewidth}
\center
 0.49
\end{minipage}
\hfill
\begin{minipage}[h]{0.085\linewidth}
\center
 0.27
\end{minipage}
\hfill
\begin{minipage}[h]{0.085\linewidth}
\center
 0.11
\end{minipage}
\hfill
\begin{minipage}[h]{0.085\linewidth}
\center
 0.01
\end{minipage}
\end{minipage}

\caption{Evolution of the matrix $\mathbf{X}^{(t)}$ with our architecture using full matrix completion model RGCNN (top) and factorized matrix completion model sRGCNN (bottom). Numbers indicate the RMS error. 
}
\label{fig:X-evolutions}
\end{figure*}

\begin{figure}[!ht]
\setlength\figureheight{5.5cm}
\setlength\figurewidth{8.5cm}
\input{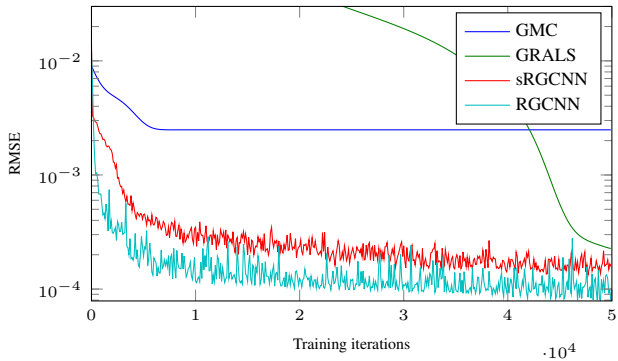}
\vspace{-5mm}
\caption{Convergence rates of the tested algorithms over the Synthetic Netflix dataset.}
\label{fig:convergence-rate}
\end{figure}

\begin{figure}[!ht]
\centering
\begin{minipage}[b]{0.2\linewidth}
\hspace*{-0.4cm}\raisebox{-0.1cm}{
\setlength\figureheight{3.2cm}
\setlength\figurewidth{3.2cm}
\begin{tikzpicture}

\begin{axis}[
xmin=0, xmax=20,
ymin=0, ymax=20,
axis on top,
width=\figurewidth,
height=\figureheight,
xtick = {0, 8.5, 17},
xticklabels = {0,1,2},
ytick = {3, 11.5, 20},
yticklabels = {2,1,0},
xticklabel pos=right,
xlabel near ticks,
axis x line = right,
axis y line = left,
y axis line style = {stealth-},
colormap={mymap}{[1pt]
  rgb(0pt)=(0,0,0.5);
  rgb(22pt)=(0,0,1);
  rgb(25pt)=(0,0,1);
  rgb(68pt)=(0,0.86,1);
  rgb(70pt)=(0,0.9,0.967741935483871);
  rgb(75pt)=(0.0806451612903226,1,0.887096774193548);
  rgb(128pt)=(0.935483870967742,1,0.0322580645161291);
  rgb(130pt)=(0.967741935483871,0.962962962962963,0);
  rgb(132pt)=(1,0.925925925925926,0);
  rgb(178pt)=(1,0.0740740740740741,0);
  rgb(182pt)=(0.909090909090909,0,0);
  rgb(200pt)=(0.5,0,0)
},
point meta min=-0.899999999999999,
point meta max=1,
colorbar style={ytick={-0.8,-0.6,-0.4,-0.2,0,0.2,0.4,0.6,0.8,1},yticklabels={−0.8,−0.6,−0.4,−0.2,0.0,0.2,0.4,0.6,0.8,1.0}}
]
\addplot graphics [includegraphics cmd=\pgfimage,xmin=0, xmax=17, ymin=20, ymax=3] {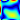};
\end{axis};
\node at (-0.1, -0.1) {\tiny $\lambda_r$};
\node at (1.65, 1.45) {\tiny $\lambda_c$};

\end{tikzpicture}
}
\end{minipage}
\hfill
\begin{minipage}[b]{0.73\linewidth}
\setlength\figureheight{2.7cm}
\setlength\figurewidth{2.98cm}
\begin{tikzpicture}
\begin{groupplot}[
     group style = {group size = 4 by 2, horizontal sep=0.1cm, vertical sep=0.1cm}]
     
\input{Figures/final_results/tikz/synthetic_netflix/2D_spectral_filters/filter_0_absolute_value/filter_.tex}
\input{Figures/final_results/tikz/synthetic_netflix/2D_spectral_filters/filter_1_absolute_value/filter_.tex}
\input{Figures/final_results/tikz/synthetic_netflix/2D_spectral_filters/filter_2_absolute_value/filter_.tex}
\input{Figures/final_results/tikz/synthetic_netflix/2D_spectral_filters/filter_3_absolute_value/filter_.tex}

\input{Figures/final_results/tikz/synthetic_netflix/2D_spectral_filters/filter_4_absolute_value/filter_.tex}
\input{Figures/final_results/tikz/synthetic_netflix/2D_spectral_filters/filter_5_absolute_value/filter_.tex}
\input{Figures/final_results/tikz/synthetic_netflix/2D_spectral_filters/filter_6_absolute_value/filter_.tex}
\input{Figures/final_results/tikz/synthetic_netflix/2D_spectral_filters/filter_7_absolute_value/filter_.tex}
  \end{groupplot}
\end{tikzpicture}
\end{minipage}
\caption{Absolute value of the first 8 spectral filters learnt by our bidimensional convolution. On the left the first filter with the reference axes associated to the row and column graph eigenvalues.}
\label{fig:MGCNN-spectral-filter}
\end{figure}

\begin{figure}[!h]
\setlength\figureheight{5.5cm}
\setlength\figurewidth{8.5cm}
\begin{tikzpicture}

\definecolor{color1}{rgb}{0.75,0,0.75}
\definecolor{color0}{rgb}{0,0.75,0.75}
\definecolor{color2}{rgb}{0.75,0.75,0}

\begin{axis}[
xlabel={$\lambda_r, \lambda_c$},
ylabel={Filter Response},
xmin=0, xmax=2,
ymin=0, ymax=0.9,
axis on top,
width=\figurewidth,
height=\figureheight
]
\addplot [blue]
table {%
0 0.136324610561132
0.1 0.210317468804121
0.2 0.228814396899939
0.3 0.205480297130346
0.4 0.152584236186743
0.5 0.0809994451701642
0.6 0.000203319591283919
0.7 0.0817225806295871
0.8 0.1580925311625
0.9 0.22361664326787
1 0.274400863796472
1.1 0.307946975189447
1.2 0.323152595478296
1.3 0.320311178284883
1.4 0.301112012821436
1.5 0.268640223890543
1.6 0.227376771885157
1.7 0.183198452788592
1.8 0.143377898174524
1.9 0.116583575206995
};
\addplot [green!50.0!black]
table {%
0 0.319112024269998
0.1 0.267679165128619
0.2 0.234495540849865
0.3 0.215986791376025
0.4 0.208853592418134
0.5 0.210071655455977
0.6 0.216891727738082
0.7 0.226839592281729
0.8 0.237716067872941
0.9 0.247597009066492
1 0.254833306185901
1.1 0.258050885323435
1.2 0.256150708340108
1.3 0.248308772865683
1.4 0.233976112298667
1.5 0.212878795806319
1.6 0.18501792832464
1.7 0.150669650558382
1.8 0.110385138981044
1.9 0.0649906058348718
};
\addplot [red]
table {%
0 0.87893633171916
0.1 0.373574490180612
0.2 0.0638341366112234
0.3 0.0931944779247045
0.4 0.136118144446611
0.5 0.0992406960576774
0.6 0.0125630079448224
0.7 0.0982170026212929
0.8 0.211704376286268
0.9 0.310807111611962
1 0.382736165076494
1.1 0.419005451074242
1.2 0.415431841915846
1.3 0.372135167828202
1.4 0.29353821695447
1.5 0.188366735354066
1.6 0.0696494270026688
1.7 0.0452820462077853
1.8 0.134793064469099
1.9 0.172946050056815
};
\addplot [color0]
table {%
0 0.289882898330688
0.1 0.0793396825909615
0.2 0.00113561244010929
0.3 0.019990401160717
0.4 0.104608240318298
0.5 0.227677799761295
0.6 0.365872227621078
0.7 0.499849150311947
0.8 0.614250672531128
0.9 0.697703377258777
1 0.74281832575798
1.1 0.746191057574749
1.2 0.708401590538025
1.3 0.634014420759678
1.4 0.531578522634507
1.5 0.413627348840237
1.6 0.296678830337525
1.7 0.201235376369953
1.8 0.151783874464035
1.9 0.17679569042921
};
\addplot [color1, dashed]
table {%
0 0.608784426935017
0.1 0.356468193937093
0.2 0.149512822747231
0.3 0.0129816011972724
0.4 0.13254165840894
0.5 0.211320595350116
0.6 0.252098324432969
0.7 0.258281424019486
0.8 0.233903138421476
0.9 0.183623377900571
1 0.112728718668223
1.1 0.0271324028857056
1.2 0.0666256613358853
1.3 0.161378899935633
1.4 0.249334072902798
1.5 0.322071274276823
1.6 0.370543932147324
1.7 0.3850788086541
1.8 0.355375999987126
1.9 0.270508936386556
};
\addplot [color2, dashed]
table {%
0 0.0335052609443665
0.1 0.0223346426725388
0.2 0.0353944054603576
0.3 0.120627729392052
0.4 0.217024522876739
0.5 0.310957327485084
0.6 0.391512032604217
0.7 0.450487875437736
0.8 0.482397441005707
0.9 0.484466662144661
1 0.456634819507599
1.1 0.401554541563988
1.2 0.324591804599762
1.3 0.233825932717324
1.4 0.140049597835541
1.5 0.056768819689751
1.6 0.000202965831756704
1.7 0.0107152483701706
1.8 0.0456602402687069
1.9 0.193688842749595
};
\addplot [black, dashed]
table {%
0 0.410960968583822
0.1 0.0308272948265077
0.2 0.217365965932607
0.3 0.360007638037205
0.4 0.420934387475252
0.5 0.421430721879005
0.6 0.380228990525007
0.7 0.313509384334087
0.8 0.234899935871363
0.9 0.155476519346237
1 0.0837628506124021
1.1 0.0257304871678354
1.2 0.0152011718451976
1.3 0.0381648856401443
1.4 0.0448455717861653
1.5 0.0394803062081337
1.6 0.028858323186636
1.7 0.0223210153579712
1.8 0.0317619337141513
1.9 0.0716267876029011
};
\addplot [blue, dashed]
table {%
0 0.292134643532336
0.1 0.00637941971868266
0.2 0.14157137349695
0.3 0.15948333171457
0.4 0.100516167972982
0.5 0.000570484437048305
0.6 0.114658232535422
0.7 0.218269923652708
0.8 0.293569645430148
0.9 0.328362725819647
1 0.316095733083785
1.1 0.255856475795806
1.2 0.152374002839625
1.3 0.016018603409827
1.4 0.137198192988336
1.5 0.285623616538941
1.6 0.401963657115399
1.7 0.45328306428045
1.8 0.401005347286165
1.9 0.200912775073947
};
\end{axis}

\end{tikzpicture}
\vspace{-2mm}
\caption{Absolute value of the first four spectral filters learned by the user (solid) and items (dashed) GCNNs. }
\label{fig:GCNN-spectral-filter}
\end{figure}
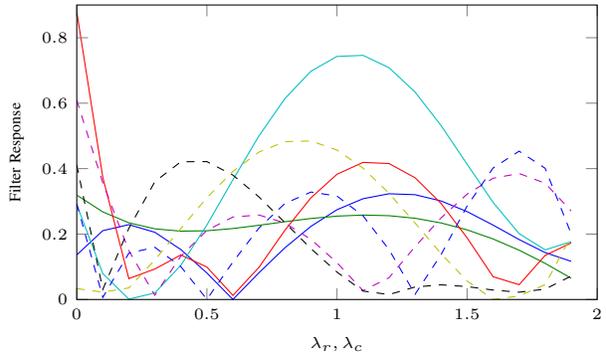

We start our experimental evaluation showing the performance of our approach on a small synthetic dataset, in which the user and item graphs have strong communities structure. Though rather simple, such a dataset allows to study the behavior of different algorithms in controlled settings. 
%
The performance of different matrix completion methods is reported in Table \ref{tab:results-synthetic-netflix}, along with their theoretical complexity. 
Our RGCNN model achieves the best accuracy, followed by the separable RGCNN. 
%
%
Different diffusion time steps of these two models are visualized in Figure \ref{fig:X-evolutions}. 
%
Figure \ref{fig:convergence-rate} shows the convergence rates of different methods. Figures \ref{fig:MGCNN-spectral-filter} and \ref{fig:GCNN-spectral-filter} depict the spectral filters learnt by the MGCNN and row- and column-GCNNs.

We repeated the same experiment considering only the column (users) graph to be given. In this setting, the RGCNN cannot be applied, while the sRGCNN has only one GCNN applied on the factor $\mathbf{H}$, and the other factor $\mathbf{W}$ is free. 
Table~\ref{tab:results-synthetic-netflix-one-graph} summarizes the results of this experiment, again, showing that our approach performs the best. 


\begin{table}[!ht]
\caption{
Comparison of different matrix completion methods using {\it users+items graphs} in terms of number of parameters (optimization variables) and computational complexity order (operations per iteration). Rightmost column shows the RMS error on Synthetic dataset. 
}
\label{tab:results-synthetic-netflix}
\vskip 0.15in
\begin{center}
\begin{small}
\begin{sc}
\begin{tabular}{lccc}
\hline
\abovespace\belowspace
Method & Parameters & Complexity & RMSE\\
\hline
\abovespace
GMC   & $m n $ & $m n$ & 0.3693   \\
GRALS & $m+n$ & $m+n$ & 0.0114 \\
{\bf RGCNN}    & $\mathbf{1}$ & $\mathbf{m n}$ & {\bf 0.0053} \\
{\bf  sRGCNN}  & $\mathbf{1}$ &$\mathbf{m+n}$ & {\bf 0.0106} \\
\hline
\end{tabular}
\end{sc}
\end{small}
\end{center}
\vskip -0.1in
\end{table}

\begin{table}[!ht]
\caption{Comparison of different matrix completion methods using {\it users graph only} in terms of number of parameters (optimization variables) and computational complexity order (operations per iteration). Rightmost column shows the RMS error on Synthetic dataset.}
\label{tab:results-synthetic-netflix-one-graph}
\vskip 0.15in
\begin{center}
\begin{small}
\begin{sc}
\begin{tabular}{lccc}
\hline
\abovespace\belowspace
Method & Parameters & Complexity & RMSE \\
\hline
\abovespace
GRALS & $m+n$ & $m+n$  & 0.0452 \\
{\bf sRGCNN} & $\mathbf{m}$ & $\mathbf{m+n}$ & {\bf 0.0362}\\
\hline
\end{tabular}
\end{sc}
\end{small}
\end{center}
\vskip -0.1in
\end{table}

\subsection{Real data}

Following \cite{rao2015collaborative}, we evaluated the proposed approach on the MovieLens, Flixster, Douban and YahooMusic datasets.  
%
For the MovieLens dataset we constructed the user and item (movie) graphs as unweighted 10-nearest neighbor graphs in the space of user and movie features, respectively. 
For Flixster, the user and item graphs were constructed from the scores of the original matrix. On this dataset, we also performed an experiment using only the users graph. 
%
%
For the Douban dataset, we used only the user graph (the provided social network of the user). 
For the YahooMusic dataset, we used only the item graph, constructed with unweighted 10-nearest neighbors in the space of item features (artists, albums, and genres). 
For the latter three datasets, we used a sub-matrix of $3000 \times 3000$ entries for evaluating the performance. 

Tables~\ref{tab:results-movielens} and~\ref{tab:other-datasets} summarize the performance of different methods. RGCNN outperforms the competitors in all the experiments. 


\begin{table}[!h]
\caption{Performance (RMS error) of different matrix completion methods on the MovieLens dataset.\vspace{-2.25mm}}
\label{tab:results-movielens}
\vskip 0.15in
\begin{center}
\begin{small}
\begin{sc}
\begin{tabular}{lc}
\hline
\abovespace\belowspace
Method & RMSE \\
\hline
\abovespace 
Global Mean & 1.154\\
User Mean & 1.063\\
Movie Mean & 1.033\\
MC \cite{candes2012exact} & 0.973\\
IMC \cite{jain2013provable,xu2013speedup} & 1.653\\
GMC  \cite{kalofolias2014matrix} & 0.996\\
GRALS \cite{rao2015collaborative} & 0.945\\
{\bf sRGCNN} & {\bf 0.929} \\
\hline
\end{tabular}
\end{sc}
\end{small}
\end{center}
\vskip -0.1in
\end{table}


\begin{table}[!h]
\caption{
Matrix completion results on several datasets (RMS error). For Douban and YahooMusic, a single graph (of users and items, respectively) was used. For Flixter, two settings are shown: users+items graphs / only users graph. \vspace{-2.25mm} 
}
\label{tab:other-datasets}
\vskip 0.15in
\begin{center}
\begin{small}
\begin{sc}
\begin{tabular}{lccc}
\hline
\abovespace\belowspace
Method & Flixster & Douban & YahooMusic \\
\hline
\abovespace 
GRALS & 1.3126 / 1.2447 & 0.8326 & 38.0423\\ 
{\bf sRGCNN} & {\bf 1.1788} / {\bf 0.9258} & {\bf 0.8012} & {\bf 22.4149} \\
\hline
\end{tabular}
\end{sc}
\end{small}
\end{center}
\end{table}

\section{Conclusion}
\label{sec:conc}

In this paper, we presented a new deep learning approach for matrix completion based on a specially designed multi-graph convolutional neural network architecture. Among the key advantages of our approach compared to traditional methods is its low computational complexity and constant number of degrees of freedom independent of the matrix size. We showed that the use of deep learning for matrix completion allows to beat current state-of-the-art recommender system methods. To our knowledge, our work is the first application of deep learning on graphs to this class of problems. We believe that it shows the potential of the nascent field of geometric deep learning on non-Euclidean domains, and will encourage future works in this direction.


\section{Acknowledgments}
FM and MB are supported in part by ERC Starting Grant No. 307047 (COMET), ERC Consolidator Grant No. 724228 (LEMAN), Google Faculty Research Award, Nvidia equipment grant, Radcliffe fellowship from Harvard Institute for Advanced Study, and TU Munich Institute for Advanced Study, funded by the German Excellence Initiative and the European Union Seventh Framework Programme under grant agreement No. 291763. XB is supported in part by NRF Fellowship NRFF2017-10.

\bibliography{sections/refsx,sections/refs,sections/refs1}
\bibliographystyle{icml2017}

\end{document}